\documentclass[final]{IEEEtran}

\usepackage[english]{babel}
\usepackage[noadjust]{cite}
\usepackage[utf8]{inputenc}

\usepackage{hyperref}
\usepackage[nolist]{acronym}
\include{notation}
\usepackage[most]{tcolorbox}

\usepackage{array}
\newcommand{\PreserveBackslash}[1]{\let\temp=\\#1\let\\=\temp}
\newcolumntype{C}[1]{>{\PreserveBackslash\centering}p{#1}}
\newcolumntype{R}[1]{>{\PreserveBackslash\raggedleft}p{#1}}
\newcolumntype{L}[1]{>{\PreserveBackslash\raggedright}p{#1}}

\usepackage{multirow}

\usepackage{makecell}

\usepackage{setspace}	

\usepackage{siunitx}
\usepackage{graphicx} 
\usepackage{epsfig} 
\usepackage{pgfplots}
\usepackage{caption}
\usepackage[update,prepend]{epstopdf}
\usepackage[font=footnotesize]{subcaption}
\usepackage{pdfpages}

\usepackage{float}

\usepackage[export]{adjustbox}

\usepackage{xcolor}
\usepackage{colortbl} 
\definecolor{myYellow}{rgb}{0.93,0.69,0.13}
\definecolor{myPurple}{rgb}{0.49,0.18,0.56}
\definecolor{myGreen}{rgb}{0.26 0.72 0.54}

\usepackage{array} 
\usepackage{wrapfig} 
\usepackage{pifont}
\usepackage{dblfloatfix}
\usepackage{placeins}
\newcommand{\cmark}{
    {\noindent\color{green}{\ding{51}}}
}%
\newcommand{\xmark}{
    {\noindent\color{red}{\ding{55}}}
}%
\newcommand{\amark}{
    {\noindent\color{orange}{\ding{81}}}
}%

\usepackage{mathtools}
\usepackage{nicefrac}
\usepackage{amsmath}
\usepackage{amssymb}
\usepackage{bm} 

\usepackage{algorithm}
\usepackage[noend]{algpseudocode}

\makeatletter
\def\BState{\State\hskip-\ALG@thistlm}
\makeatother

\usepackage{tikz}
\pgfdeclarelayer{foreground}
\pgfsetlayers{background, main, foreground}
\usetikzlibrary{quotes, angles, backgrounds, arrows, automata, shapes, positioning, calc, through, spy, decorations.pathreplacing, decorations.markings, arrows.meta, automata, petri}

\tikzset{
    imglabel/.style={
      rectangle,
      inner sep=2pt,
      text=black,
      minimum height=1em,
      text centered,
      fill=white,
      fill opacity=1.0,
      text opacity=1,
      anchor=south west,
    },
  }
\tikzset{
    arrow at end/.style={
        decorate,decoration={
            markings,
            mark=at position .999 with{
                \arrow{#1};
            }
        }
    }
}
\tikzset{
	state/.style={
		rectangle,
		draw=black, very thick,
		minimum height=1.0em,
		text centered,
	},
}
\tikzset{
  on each segment/.style={
    decorate,
    decoration={
      show path construction,
      moveto code={},
      lineto code={
        \path [#1]
        (\tikzinputsegmentfirst) -- (\tikzinputsegmentlast);
      },
      curveto code={
        \path [#1] (\tikzinputsegmentfirst)
        .. controls
        (\tikzinputsegmentsupporta) and (\tikzinputsegmentsupportb)
        ..
        (\tikzinputsegmentlast);
      },
      closepath code={
        \path [#1]
        (\tikzinputsegmentfirst) -- (\tikzinputsegmentlast);
      },
    },
  },
  mid arrow/.style={postaction={decorate,decoration={
        markings,
        mark=at position .5 with {\arrow[#1]{stealth}}
      }}},
}

\graphicspath{{./figures/}}

\newcommand\copyrighttext{%
    \small \begin{center} \color{red} \textcopyright\,2024 IEEE. Accepted for IEEE Communications Magazine. Personal use of this material is permitted. Permission from IEEE must be obtained for all other uses, in any current or future media, including reprinting/republishing this material for advertising or promotional purposes, creating new collective works, for resale or redistribution to servers or lists, or reuse of any copyrighted component of this work. \end{center}}
\newcommand\copyrightnotice{%
	\begin{tikzpicture}[remember picture,overlay]
	\node[anchor=south,yshift=25.6cm] at (current page.south) 
	{\color{red}\fbox{\parbox{\dimexpr\textwidth-\fboxsep-\fboxrule\relax}{\copyrighttext}}};
	\end{tikzpicture}%
}

\begin{document}

\title{\copyrightnotice Reshaping UAV-Enabled Communications with Omnidirectional Multi-Rotor Aerial Vehicles
}

\author{Daniel Bonilla Licea$^{1,2}$, Giuseppe Silano$^{2}$, Hajar El Hammouti$^{1}$, Mounir Ghogho$^{3}$, and Martin Saska$^{2}$
    \thanks{$^1$D. Bonilla Licea and H. El Hammouti are with the College of Computing, Mohammed VI Polytechnic University, Ben Guerir, Morocco (emails: {\tt\small \{daniel.bonilla, hajar.elhammouti\}@um6p.ma}).}
    \thanks{$^2$D. Bonilla Licea, G. Silano, and M. Saska are with the Czech Technical University in Prague, Prague, Czech Republic (emails: {\tt\small \{bonildan, giuseppe.silano, martin.saska\}@fel.cvut.cz}).}
    \thanks{$^{3}$M. Ghogho is with the International University of Rabat, Rabat, Morocco (email: {\tt\small mounir.ghogho@uir.ac.ma}).}
    \thanks{This work was partially funded by EU under ROBOPROX (reg. no. CZ.02.01.01/00/22\_008/0004590), by EU under the ARISE programme grant no. DCI-PANAF/2020/420-02, by the Czech Science Foundation (GAČR) project no. 23-07517S, and by the CTU grant no. SGS23/177/OHK3/3T/13.}
}

\maketitle


\begin{acronym}
    \acro{AoA}[AoA]{Angle of Arrival}
    \acro{AoD}[AoD]{Angle of Departure}
    \acro{BS}[BS]{Base Station}
    \acro{CoM}[CoM]{Center of Mass}
    \acro{DoF}[DoF]{Degrees of Freedom}
    \acro{FSO}[FSO]{Free-Space Optical}
    \acro{IoT}[IoT]{Internet of Thing}
    \acro{LoS}[LoS]{Line-of-Sight}
    \acro{MRAV}[MRAV]{Multi-Rotor Aerial Vehicle}
    \acro{UAV}[UAV]{Unmanned Aerial Vehicle}
    \acro{f-MRAV}[f-MRAV]{fully actuated MRAV}
    \acro{u-MRAV}[u-MRAV]{under-actuated MRAV}
    \acro{o-MRAV}[o-MRAV]{omnidirectional MRAV}
    \acro{RF}[RF]{Radio Frequency}
    \acro{RSSI}{Received Signal strength Indicator}
    \acro{SNR}[SNR]{Signal-to-Noise Ratio}
    \acro{SINR}[SINR]{Signal-to-Interference-plus-Noise Ratio}
    \acro{wrt}[w.r.t.]{with respect to}
\end{acronym}



\begin{abstract}
A new class of \acp{MRAV}, known as \acp{o-MRAV}, has attracted significant interest in the robotics community. These \acp{MRAV} have the unique capability of independently controlling their 3D position and 3D orientation. In the context of aerial communication networks, this translates into the ability to control the position and orientation of the antenna mounted on the \ac{MRAV} without any additional devices tasked for antenna orientation. This additional \ac{DoF} adds a new dimension to aerial communication systems, creating various research opportunities in communications-aware trajectory planning and positioning. This paper presents this new class of \acp{MRAV} and discusses use cases in areas such as physical layer security and optical communications. Furthermore, the benefits of these \acp{MRAV} are illustrated with realistic simulation scenarios. Finally, new research problems and opportunities introduced by this advanced robotics technology are discussed.
\end{abstract}



\begin{IEEEkeywords}
    Multi-Rotor Aerial Vehicles, Communication-Aware Robotics, Pose Optimization, Antenna Orientation.
\end{IEEEkeywords}



\section{Introduction} 
\label{sec:intro}

Over the past decade, the integration of \acp{UAV} into wireless networks has attracted significant attention from both academic and industrial research communities \cite{MozaffariIEEECST2019}. \acp{UAV} have been extensively studied for two main reasons: their agility allows for swift deployment to address changing demands and optimize network performance \cite{MozaffariIEEECST2019}, 
and their ability to achieve \ac{LoS} improves communication reliability and expands coverage area \cite{BonillaPIEEE2024}. 
Consequently, \acp{UAV} have been considered for applications such as replacing damaged base stations in disaster recovery, serving as data relays for remote \ac{IoT} devices, and acting as aerial base stations to enhance network capacity in urban and underserved areas \cite{BonillaPIEEE2024, MozaffariIEEECST2019}.

Much of the literature focuses on \acp{u-MRAV}, which have fewer control inputs -- such as thrust, pitch, roll, and yaw rate -- than \acf{DoF} in 3D space. This means that the 3D orientation of \acp{u-MRAV} is dependent on their translational velocity, preventing independent control of position and orientation \cite{HamandiIJRR2021, Hamandi2021AIRPHARO}. Despite their mechanical simplicity and reduced cost, \acp{u-MRAV} face challenges such as tilting during motion or under external forces (e.g., wind), complicating communication-aware trajectory planning \cite{Muralidharan2021ARCRAS}. This issue is further exacerbated with high-frequency technologies like terahertz and mmWave, which require precise antenna alignment due to high directivity \cite{BonillaPIEEE2024}. Many studies incorrectly assume that \acp{u-MRAV} can maintain a stable 3D orientation while in motion, which is only true at low velocities and small pitch/roll angles \cite{HamandiIJRR2021}.

Recent works work has explored using servo motors or other mechanisms to adjust antenna orientation on \acp{u-MRAV}. For example, \cite{KoruASME2024ToM} proposed a control law for dynamically adjusting antenna orientation, while \cite{YuCCNC2023} introduced a full-duplex communication system to manage interference through \ac{UAV}-relative position control. Another study, \cite{LiIET2019}, used reinforcement learning for antenna alignment. However, these solutions focus on highly directional antennas for one-to-one communication.

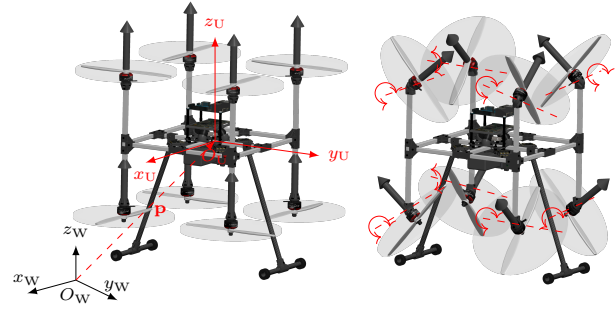
\begin{figure}[t]
    \centering
    \scalebox{0.925}{
    \begin{tikzpicture}
        \node at (0,0) [text centered] {\adjincludegraphics[trim={{.0\width} {.0\height} {.0\width} {.0\height}}, clip, width=0.9\columnwidth]{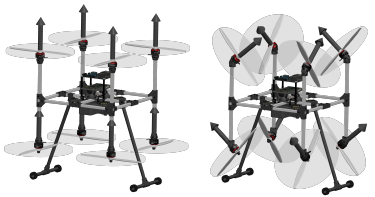}};

        \draw[-latex,red] (-1.85,0) -- (-1.85,1.5) node[above]{{\scriptsize{$z_\mathrm{U}$}}}; 
        \draw[-latex,red] (-1.85,0) -- (-0.35,-0.21) node[right]{{\scriptsize{$y_\mathrm{U}$}}}; 
        \draw[-latex,red] (-1.85,0) -- (-2.85,-0.31) node[below,red]{{\scriptsize{$x_\mathrm{U}$}}}; 
        \node at (-1.85,0) [below, red]{{\scriptsize{$O_\mathrm{U}$}}};

        \draw[-latex] (-3.85,-2) -- (-3.85,-1.5) node[above]{{\scriptsize{$z_\mathrm{W}$}}}; 
        \draw[-latex] (-3.85,-2) -- (-3.25,-2.29) node[above]{{\scriptsize{$y_\mathrm{W}$}}}; 
        \draw[-latex] (-3.85,-2) -- (-4.55,-2.19) node[above]{{\scriptsize{$x_\mathrm{W}$}}}; 
        \node at (-3.85,-2) [below]{{\scriptsize{$O_\mathrm{W}$}}};

        \draw[dashed, red] (0.60,0.66) -- (1.45,0.96); 
        \draw[<->, red]  (0.75,0.69) arc[start angle=0,end angle=280,radius=0.15];
        \draw[dashed, red] (1.2,1.16) -- (2.45,0.96); 
        \draw[<->, red]  (1.25,1.28) arc[start angle=90,end angle=-70,radius=0.15];
        \draw[dashed, red] (2.00,0.81) -- (3.15,0.31); 
        \draw[<->, red]  (2.20,0.78) arc[start angle=0,end angle=280,radius=0.15];
        \draw[dashed, red] (2.90,0.66) -- (3.70,1.01); 
        \draw[<->, red]  (3.80,0.96) arc[start angle=0,end angle=280,radius=0.15];
        \draw[dashed, red] (0.45,-1.26) -- (1.45,-0.66); 
        \draw[<->, red]  (0.66,-1.23) arc[start angle=0,end angle=280,radius=0.15];
        \draw[dashed, red] (1.25,-0.51) -- (2.35,-0.81); 
        \draw[<->, red]  (1.25,-0.41) arc[start angle=90,end angle=-70,radius=0.15];
        \draw[dashed, red] (2.00,-1.11) -- (3.15,-1.31); 
        \draw[<->, red]  (2.20,-1.11) arc[start angle=0,end angle=280,radius=0.15];
        \draw[dashed, red] (2.95,-1.11) -- (3.60,-0.81); 
        \draw[<->, red]  (3.15,-1.11) arc[start angle=0,end angle=240,radius=0.15];

        \draw[-latex, dashed,red] (-3.85,-2) -- node[right]{{\scriptsize{$\mathbf{p}$}}}(-1.85,0);
    \end{tikzpicture}
    }
    \caption{Illustration of \ac{u-MRAV} and \ac{o-MRAV} configurations with the global and untilted reference frames \cite{Aboudorra2023JINT}. Arcs indicate servo rotation for thrust vector adjustment.}
    \label{fig:mrav}
\end{figure}

The robotics community has introduced \acp{f-MRAV} and \acfp{o-MRAV} \cite{HamandiIJRR2021, Hamandi2021AIRPHARO}. These designs offer more control inputs than \ac{DoF}, allowing for partial (\acp{f-MRAV}) or full (\acp{o-MRAV}) independent control over all movement axes. This capability greatly expands their maneuverability and allows precise antenna control throughout the \ac{UAV}'s mission. From a communication standpoint, this feature is crucial for optimizing both the 3D position and orientation of the antenna. For instance, \acp{o-MRAV} with dipole antennas can maintain communication with multiple nodes simultaneously, providing more flexibility in dynamic environments. Figure \ref{fig:mrav} presents an illustrative example of both an \ac{u-MRAV} and an \ac{o-MRAV}.

While \acp{o-MRAV} introduce additional complexity due to their design and control requirements, they offer significant advantages in wireless networks by providing highly adaptive aerial platforms~\cite{BonillaICASSP2024}. Drawing on the authors' previous work \cite{BonillaICASSP2024, BonillaPIEEE2024, Bonilla2024Globecom}, this paper explores how the independent control of both position and orientation in \acp{o-MRAV} can enhance communication systems, particularly in areas like physical layer security and beamforming.



\section{\acp{o-MRAV}: Design, Modeling and Control}
\label{sec:o-MRAV}

When designing \acp{o-MRAV}, several actuation strategies have been proposed in the literature. According to \cite{HamandiIJRR2021, Aboudorra2023JINT}, \acp{o-MRAV} can be categorized into three classes: (i) \textit{bi-directional} propellers, which reverse rotation to generate thrust in both directions, (ii) \textit{uni-directional} propellers, generating thrust in one direction (clockwise or counterclockwise), and (iii) \textit{actively tilted} propellers, which individually tilt around their axes using servo motors. The third class is the most flexible, allowing for optimized propeller orientation and rotation to improve platform efficiency and payload capacity, though it introduces complexity in mechanical design and control.

In this study, an actively titled \ac{o-MRAV} with eight bi-directional propellers is considered. These propellers are attached to the vertices of a cube centered at the vehicle's \ac{CoM}, as discussed in \cite{Aboudorra2023JINT} and depicted in Figure \ref{fig:mrav}. Each propeller tilts around a different axis by the same angle, referred to as the \textit{tilting angle}. This configuration allows the total thrust vector to point in all three orthogonal axis directions.

Modeling the \ac{o-MRAV} involves defining two reference frames: the untilted frame, centered at the \ac{CoM}, and the global frame (see Figure \ref{fig:mrav}). The platform's dynamics are modeled as a six-\ac{DoF} floating rigid body using Newton-Euler formalism, with the state including position, orientation, velocity, and angular velocity. Control inputs consist of the rotor speeds, spinning directions, and tilt angles of the eight bi-directional propellers. Forces and torques generated by the rotors at the \ac{CoM} depend on rotor positions and orientation. More details are provided in \cite{HamandiIJRR2021, Hamandi2021AIRPHARO, Aboudorra2023JINT}.



\subsection{Comparative evaluation of \acp{o-MRAV} capabilities} %
\label{sec:systemAbilities}

The capabilities of \acp{o-MRAV} are compared to their under-actuated (\acp{u-MRAV}) and fully actuated (\acp{f-MRAV}) counterparts. As detailed in \cite{HamandiIJRR2021, Aboudorra2023JINT}, these capabilities are closely tied to the mechanical design of each platform. While \acp{f-MRAV} offer partial independent control of position and orientation, they are limited in many directions. In contrast, \acp{o-MRAV} provide full independent control over both 3D position and 3D orientation, enabling more precise and complex maneuvers. This is due to the ability of \acp{o-MRAV} to control both the spinning direction and speed of rotors, along with propeller tilting angles (see Figure \ref{fig:mrav}). \acp{f-MRAV}, by comparison, have fixed tilting angles and, in most cases, fixed spinning directions, with only motor speed as a variable. For a thorough comparison, three representative features are considered: hovering ability, trajectory tracking, and rotor failure robustness.

\textbf{Hovering ability}. The \textit{ability to hover} refers to the capability of \acp{MRAV} to remain stationary at a desired position, crucial for applications like aerial \acp{BS} or as communication relays \cite{BonillaPIEEE2024}, where multi-rotors offer a clear advantage over fixed-wing \acp{UAV}. As outlined in \cite{HamandiIJRR2021, Aboudorra2023JINT}, hovering can be classified as static or dynamic. \textit{Static hovering} allows the platform to stabilize both position and orientation with zero linear and angular velocity, enabling heading adjustments. \textit{Dynamic hovering} involves maintaining position with controlled movement and velocity around a point.

\acp{u-MRAV} can achieve static hovering and, in some cases, dynamic hovering, with uncontrolled orientation variations (e.g., rotation around a point after a propeller failure). However, this dynamic hovering is typically limited to safety or recovery scenarios and is less energy-efficient \cite{BonillaICASSP2024}. In contrast, \acp{f-MRAV} and \acp{o-MRAV} can perform both static and dynamic hovering with fully controlled orientation. The main distinction is that \acp{o-MRAV} are capable of maintaining a constant position while achieving full control of their orientation in any direction in space. This constitutes the main advantage of \acp{o-MRAV} compared to both \ac{u-MRAV} and \ac{f-MRAV} designs. This is especially useful for tasks like communication relay, where a rigidly attached antenna can be aimed at a target without moving the vehicle, resulting in more efficient operations. 

\textbf{Trajectory tracking ability}. The ability to follow a trajectory is crucial for many applications and can be classified based on the type (position and orientation tracking) and the number of \ac{DoF} the \ac{MRAV} can independently control. For standard \acp{u-MRAV}, only three-dimensional position and one-dimensional orientation tracking are possible, as their pitch and roll angles are coupled with translational movements, limiting control over these axes. Some designs, such as those with radially tilted or tilting propellers \cite{HamandiIJRR2021}, extend this capability to achieve more \ac{DoF}, moving these systems from under-actuated to over-actuated, thus classifying them as \acp{f-MRAV} and \acp{o-MRAV}.
Most \acp{f-MRAV} can track three-dimensional position and two-dimensional orientation, and some advanced designs can achieve full three-dimensional attitude tracking \cite{HamandiIJRR2021, Lei2024TAES}. However, these platforms still cannot fully exploit their \ac{DoF}, limiting their applications despite some advantages over \acp{u-MRAV}. Only \acp{o-MRAV} can achieve independent tracking of both three-dimensional position and orientation, making them particularly effective in windy or turbulent conditions, reducing the jittering effects commonly seen in \acp{u-MRAV} \cite{WangIEEETWC2022}.

\textbf{Rotor failure robustness}. While \acp{f-MRAV} and \acp{o-MRAV} are commonly believed to be more resilient to rotor failures than \acp{u-MRAV} due to their higher number of rotors\footnote{Typically, \acp{u-MRAV} have four to eight rotors, often with opposed propellers to enhance lift \cite{HamandiIJRR2021}.}, their true robustness lies in their ability to perform controlled hovering, both static and dynamic. In the event of rotor failure, where a propeller becomes unusable and ceases to generate thrust, \acp{u-MRAV} struggle to maintain attitude control, making safe landing challenging unless specialized algorithms or additional sensors are employed. Although \acp{u-MRAV} may achieve dynamic hovering by rotating uncontrollably around a fixed point in failure scenarios, this is primarily for recovery and safety, and it is not energy-efficient for long-term use. 

By contrast, \acp{f-MRAV} and \acp{o-MRAV} are capable of maintaining stable flight even after losing one or more rotors, thanks to their more advanced control mechanisms \cite{Hamandi2021AIRPHARO}. Specifically, \acp{o-MRAV}, with their ability to independently reorient propellers and dynamically adjust spinning directions, exhibit superior flexibility and robustness. This allows them to adapt more effectively to rotor failures compared to \acp{f-MRAV}, ensuring continued operation in more challenging conditions.

Table \ref{tab:MARV-comparison} compares the capabilities of different \ac{MRAV} designs, categorizing their abilities as ``included'' ({\cmark}), ``partially included'' ({\amark}), or ``not included'' ({\xmark}), to highlight the differences between \ac{u-MRAV}, \ac{f-MRAV}, and \ac{o-MRAV} platforms.

\begin{table}[t]
    \centering
    \caption{Comparison of included system abilities, based on the \ac{MRAV} design: ``included'' ({\cmark}), ``partially included'' ({\amark}), and ``not included'' ({\xmark})}
    \label{tab:MARV-comparison}
    \vspace*{-4.0mm}
    \begin{center}
    \renewcommand{\arraystretch}{1.1}
    \setlength{\tabcolsep}{5pt}
    \scalebox{0.8}{
    \begin{tabular}{c | cccccc }
    \hline
        \multirow{3}{1.45cm}{\centering\textbf{\ac{MRAV} designs}} & \multicolumn{5}{c}{\textbf{System Abilities}} \\
        & \multirow{2}{1.45cm}{\centering Static Hovering} & \multirow{2}{1.45cm}{\centering Dynamic Hovering} & \multirow{2}{1.45cm}{\centering Position Tracking} & \multirow{2}{1.45cm}{\centering Orientation Tracking} & \multirow{2}{1.45cm}{\centering Failure Robustness}  \\ \\
        \hline \hline
        \ac{u-MRAV} & \cmark & \amark & \cmark & \xmark & \cmark \\
        \ac{f-MRAV} & \cmark & \amark & \cmark & \amark & \amark \\
        \ac{o-MRAV} & \cmark & \cmark & \cmark & \cmark & \cmark \\
    \arrayrulecolor{black}\hline
    \end{tabular}
    }
    \end{center}
\end{table}



\subsection{Challenges and open problems}
\label{sec:challengesOpenProblems}

The development and deployment of \acp{o-MRAV} pose several challenges, particularly in control complexity, energy consumption, and design costs. This section highlights these key issues and ongoing research efforts to address them.

\textbf{Control system complexity}. \acp{o-MRAV} have inherently more complex control systems due to their ability to independently manage both position and orientation. This adds complexity to 3D positioning and trajectory planning. Specifically, optimizing the orientation of antennas on \acp{o-MRAV} involves handling nonlinear, non-convex, and non-smooth optimization problems. These challenges make it difficult to find optimal solutions within a reasonable time frame, as solvers can easily become trapped in local optima. Consequently, sophisticated algorithms are required to optimize both the positioning and orientation of \acp{o-MRAV}, ensuring their efficient deployment \cite{BonillaPIEEE2024, MozaffariIEEECST2019, Muralidharan2021ARCRAS}.

\textbf{Energy consumption}. A significant challenge in the development and use of \acp{o-MRAV} is balancing performance and efficiency. For \acp{o-MRAV}, the propellers must generate sufficient lift so that the total thrust vector can overcome gravity, along with an additional margin to support dynamic maneuvers. However, one of the largest sources of inefficiency in \acp{o-MRAV} stems from the fact that some of the forces generated by the propellers are canceled out, as not all of them are aligned with the direction of motion. This occurs because opposing thrusts are required to stabilize and maneuver the vehicle, leading to wasted energy. This performance requirement is at odds with the goal of achieving high efficiency and extended flight times. Efficiency is further compromised when additional weight is introduced for actuation mechanisms. To address this challenge, ongoing research is focused on finding optimal design strategies that efficiently manage the lift produced by the propellers by adjusting the tilting angles and the spinning direction, minimizing the energy lost through opposing thrusts \cite{Aboudorra2023JINT}.

\textbf{Costs and design complexity}. Despite their advantages, \acp{o-MRAV} also have drawbacks, such as additional actuation mass, increased design complexity, and the presence of singularity cases not encountered in \acp{u-MRAV} \cite{HamandiIJRR2021, Hamandi2021AIRPHARO}. Therefore, morphology design is crucial to ensure that the resulting platform meets performance requirements. This necessitates the use of proper optimization tools\footnote{\url{https://github.com/ethz-asl/tiltrotor_morphology_optimization}} capable of performing parametric optimization of the tilting angles and the propeller positions relative to the \ac{MRAV}'s \ac{CoM}. The goal is to maximize the force and torque envelope and their maximum values in each axis direction. As a result, compared to their more commonly used \ac{u-MRAV} counterparts, many of \ac{o-MRAV} platforms are designed and developed within laboratory environments, increasing their costs \cite{Allenspach2020IJRR}.



\section{Use Cases}
\label{sec:useCases}

\begin{figure*}
    \centering
    \includegraphics[scale=0.55]{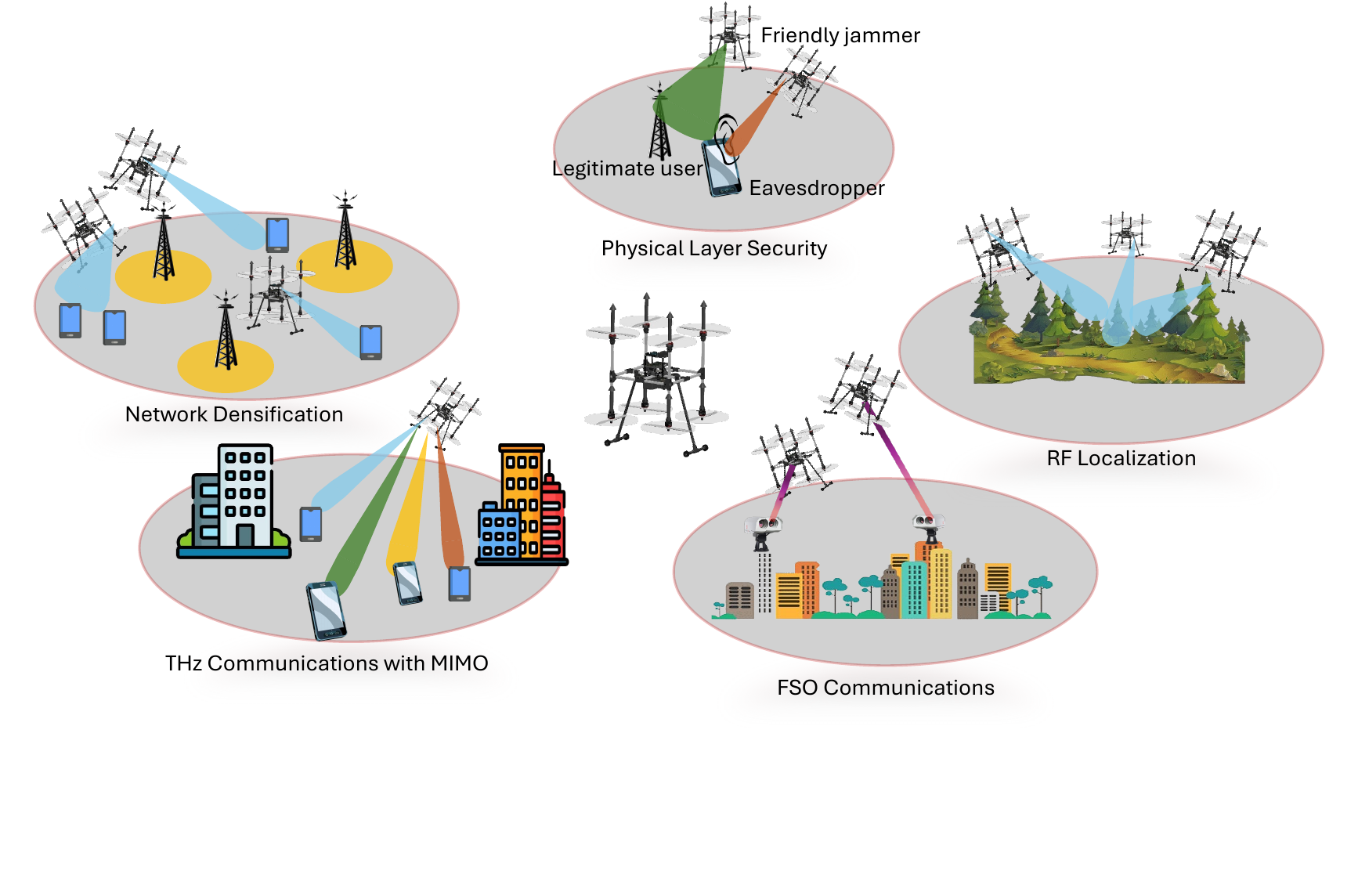}
    \vspace*{-2.25cm}
    \caption{\ac{o-MRAV} use cases.}
    \label{fig:usecases}
    \vspace*{-1em}
\end{figure*}

The primary advantage of \acp{o-MRAV} is their ability to independently control both orientation and position. From a communication perspective, this capability means that, when the antenna is fixed to the \acp{o-MRAV} frame, both its 3D position and 3D orientation can be precisely managed. This unlocks new and interesting opportunities in the field. 
This section outlines potential use cases in wireless communication networks where the unique characteristics of \acp{o-MRAV} offer significant enhancements and introduce innovative solutions to traditional challenges. Figure \ref{fig:usecases} summarizes the studied use cases.

\subsection{Physical layer security}
\label{sec:phisicalLayerSecurity}

A traditional challenge in physical layer security is maximizing the secrecy rate during communication between two legitimate nodes in the presence of an eavesdropper. Conventional strategies often employ multi-antenna arrays to direct the radiation pattern's null towards the eavesdropper while enhancing the antenna gain experienced by the legitimate receiver. 

However, if the legitimate transmitter is an \ac{o-MRAV}, similar results can be achieved using a single antenna. This is accomplished by manipulating the 3D orientation of the \ac{o-MRAV}, allowing the null of the \ac{o-MRAV} antenna's radiation pattern to be directed towards the eavesdropper while maintaining communication with the legitimate node. 

In a related scenario, if a legitimate network faces an attack by a malicious jammer, and the legitimate receivers are \acp{o-MRAV} with single antennas, they can reorient themselves to direct their antenna radiation pattern's null towards the jammer. This strategic repositioning can mitigate and neutralize the jamming attack \cite{BonillaICASSP2024}. 



\subsection{Localization of RF sources}
\label{sec:localizationRFsources}

Localization of \ac{RF} sources is a critical task in various applications, such as search and rescue operations, and identifying the position of malicious jammers. \acp{MRAV} equipped with directional antennas and sometimes additional mechanical systems for antenna rotation are commonly used for this task. Typically, \acp{MRAV} vary their yaw angle to orient the antenna in different directions and collect data to localize the target. While effective, this method has limitations in terms of flexibility and speed.

In contrast, \acp{o-MRAV} offer significant advantages in \ac{RF} source localization. These advanced \acp{MRAV} can independently control the roll and pitch angles, in addition to the yaw angle. This enhanced maneuverability allows for the design of more efficient trajectories and enable faster and more accurate localization of \ac{RF} sources even without a special system for antenna positioning. By leveraging their full range of motion, \acp{o-MRAV} achieve superior performance in tasks requiring precise and rapid localization of \ac{RF} signals. 



\subsection{Aerial \acl{FSO} communications}
\label{sec:aerialFSOCommunications}

\ac{FSO} communications is one of the most promising applications for \acp{o-MRAV}. Achieving successful \ac{FSO} communication requires establishing direct links between the transmitter and receiver, particularly when using flying robots. This leads to the alignment problem \cite{DabiriIEEETVT2019}. The transmitting \acp{MRAV} must precisely direct its laser beam to the center of the optical receiver on the other \acp{MRAV}, while the receiving \acp{MRAV} must align its lens and photodiode with the incoming laser beam. Maintaining a precise alignment is crucial for effective \ac{FSO} communication, as even slight deviations can lead to signal loss or degradation.

\acp{u-MRAV} often struggle with this alignment due to their limited ability to control orientation independently of their position. In contrast, \acp{o-MRAV} offer a substantial advantage in this regard. Their flexible orientation capabilities allow for precise positioning and alignment of the laser beam and optical receiver.

Additionally, \acp{u-MRAV} are more prone to jittering, which further complicates the alignment problem. Using \acp{o-MRAV} mitigates the impact of jittering, as these \acp{MRAV} are less sensitive to such disturbances.



\subsection{THz and mmWave communications}
\label{sec:thzMmWaveCommunications}

Some of the key technologies expected to be deployed in 6G networks include terahertz communications and mmWave communications. The extremely high frequencies of these systems will provide very large bandwidths but will also introduce significant absorption losses. To compensate for these losses, 6G systems will utilize high-gain antennas which are highly directional.

When establishing aerial links between \acp{MRAV} using these technologies, the same alignment problem described for aerial \ac{FSO} communications will arise (see Section \ref{sec:aerialFSOCommunications}). High directivity antennas require precise alignment to maintain a strong communication link. Consequently, \acp{o-MRAV} will exhibit significant advantages when establishing aerial \ac{RF} links using these technologies. Their ability to independently control position and orientation will simplify the alignment process, enhance link stability, and improve the overall reliability of 6G communications involving \acp{MRAV}.



\subsection{Capacity enhancement in dense areas}
\label{sec:capacityEnhancement}

The goal of network densification is to enhance network performance and coverage by deploying a large number of small cells within a target area. \acp{MRAV} have been considered as rapidly deployable aerial \acp{BS} to improve network quality of service. However, network densification can result in high interference and increased coordination costs, leading to additional optimization and planning challenges.

To address this challenge, \acp{o-MRAV} can mechanically control the antenna's orientation and direct the signal to reduce interference. By equipping an \ac{o-MRAV} with a highly directional antenna and combining it with an effective orientation and position controller, communication can be provided in a highly focused and dynamic manner. This approach allows connectivity to be delivered to specific users, thereby improving the spatial sharing of the electromagnetic spectrum and minimizing interference \cite{MozaffariIEEECST2019}.



\section{Simulation Experiments: \acp{o-MRAV} for Jamming and Eavesdropping Mitigation}
\label{sec:simulationExperiments}

This section examines two scenarios in the area of physical layer security where \acp{o-MRAV} demonstrate significant advantages over \acp{u-MRAV}. The first scenario involves an \ac{o-MRAV} communicating with a legitimate node while countering a jammer, showing how the \ac{o-MRAV} can optimize antenna direction to enhance communication efficiency \cite{BonillaICASSP2024}. The second scenario addresses eavesdroppers attempting to intercept communication, where a friendly jammer is used to disrupt eavesdropping and safeguard the transmission \cite{Bonilla2024Globecom}.

While this study focuses on comparing \acp{o-MRAV} and \acp{u-MRAV} due to the pronounced mechanical and operational differences between these platforms, it is worth noting that comparing \acp{o-MRAV} with \acp{f-MRAV} could provide additional insights, particularly for applications that require a higher degree of orientation control. However, \acp{f-MRAV} still lack the full 3D orientation control offered by \acp{o-MRAV}, as discussed in Section \ref{sec:systemAbilities}, which is a critical factor in the communication scenarios considered here. As such, the comparison has been prioritized between \acp{o-MRAV} and \acp{u-MRAV}, where the differences in capabilities are more pronounced. Future work could explore a detailed comparison with \acp{f-MRAV}, especially for use cases where partial orientation control may be sufficient.

\begin{figure}[h]
    \centering
    \includegraphics[scale=0.48]{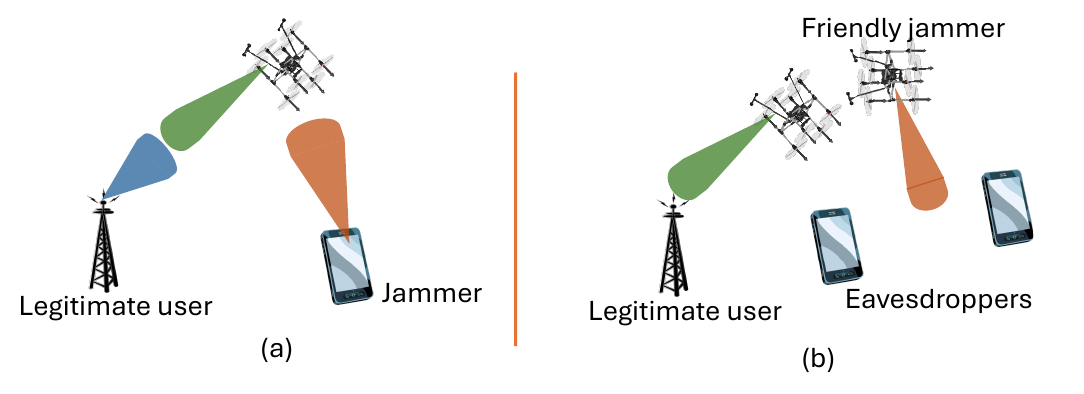}
    \vspace*{-0.75em}
    \caption{(a) An \ac{o-MRAV} communicating with a legitimate user in the presence of a jammer, (b) a friendly jammer disrupting eavesdroppers.}
    \label{fig:scenarios}
\end{figure}



\subsection{Enhancing security against jamming with \acp{o-MRAV}}
\label{sec:jammer}

A legitimate network consisting of two ground nodes and one \ac{o-MRAV} is considered, as shown in Figure \ref{fig:scenarios}(a) and detailed in \cite{BonillaICASSP2024}. The \ac{o-MRAV} receives data streams from the ground nodes while a malicious ground node transmits a jamming signal aimed at disrupting communications. The goal is to optimize the 3D position and orientation of the \ac{o-MRAV} to maximize the minimum \ac{SINR} for both ground users.

Four different strategies are tested to achieve this: (i) \textit{Optimum Pose}, numerically optimizing the position and orientation of the \ac{o-MRAV}; (ii) \textit{Maximum Gain}, orienting the \ac{o-MRAV} so that the main lobe of its antenna radiation pattern points towards the ground users, then optimizing its position numerically; (iii) \textit{Zero Interference}, orienting the \ac{o-MRAV} so that the null of its antenna radiation pattern points towards the jammer, followed by numerical optimization of its position; (iv) \textit{Vertical Orientation}, optimizing the position of an \ac{u-MRAV} with a fixed, vertically oriented antenna.

\begin{figure}[tb]
    \centering
    \adjincludegraphics[trim={{.05\width} {.225\height} {.05\width} {.225\height}},clip,width=0.95\columnwidth]{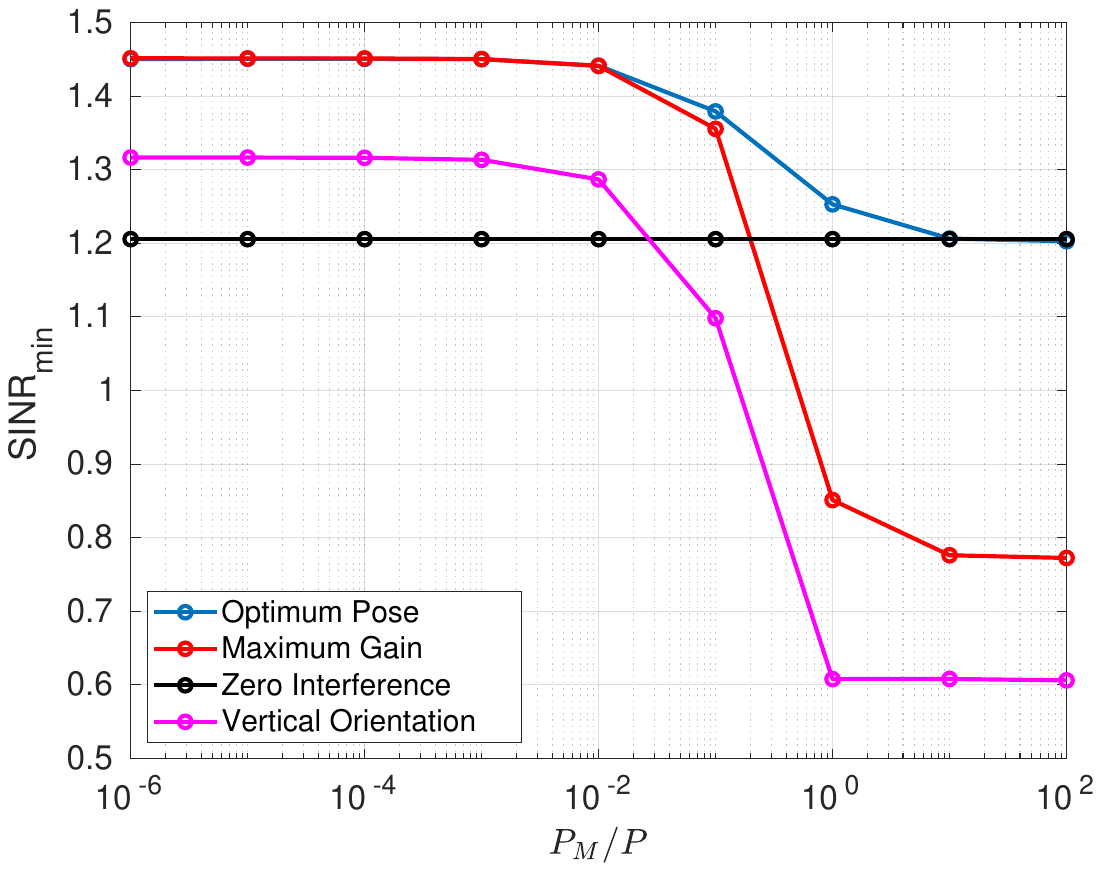}
    \caption{Minimum \ac{SINR} for different jamming powers.}
    \label{fig:2}
\end{figure}

From the results presented in Figure \ref{fig:2}, several observations can be made. When the jamming signal is weak, the \textit{Maximum Gain} strategy aligns with the optimal solution. As the jamming signal becomes stronger, the \textit{Zero Interference} strategy becomes optimal indicating that suboptimal solutions can achieve optimal poses under extreme jamming conditions. 

Another observation is that all solutions saturate as the jamming signal strength increases. The performance of the \textit{Zero Interference} strategy remains stable because it consistently nullifies the jamming signal. The optimal solution also converges towards the \textit{Zero Interference} strategy, effectively neutralizing the jammer. In contrast, the \textit{Maximum Gain} strategy, while maximizing antenna gain towards the legitimate users, it gradually adjusts its position so that the null of its antenna points towards the jammer. The \textit{Vertical Orientation} strategy positions the \ac{u-MRAV} directly above the jammer, ensuring that the null of the antenna points towards it, thereby neutralizing the jamming effect. These results demonstrate the clear advantages of optimizing both the position and orientation of the \ac{o-MRAV} in mitigating jamming attacks and establishing robust communication links. The flexibility of the \ac{o-MRAV} allows for effective management of the interference generated by the jammer, especially under strong jamming conditions. Further details can be found in \cite{BonillaICASSP2024}.

A comparison between the optimal performance of the \ac{o-MRAV} (blue plot in Figure \ref{fig:2}) and the \ac{u-MRAV} (magenta plot in Figure \ref{fig:2}) reveals significant differences, especially under strong jamming conditions. The \ac{o-MRAV}'s independent orientation control leads to a notable improvement in \ac{SINR}, particularly as jamming power intensifies. It is important to note that the frame of the \ac{MRAV} has the potential to modify the radiation pattern of the antenna, and this factor should be considered when optimizing the \ac{o-MRAV}'s pose.

\subsection{Friendly jammer \ac{o-MRAV} against eavesdroppers}
\label{sec:eavesdropper}

In this second scenario, a downlink communication is considered where an \ac{o-MRAV} communicates with a legitimate ground node while two eavesdroppers attempt to intercept the transmission, as depicted in Figure \ref{fig:scenarios}(b) and described in \cite{Bonilla2024Globecom}. To secure the communication, a second \ac{o-MRAV} is deployed to jam the eavesdroppers. The objective is to optimize the position, orientation, and transmission power of both \acp{o-MRAV} to maximize the secrecy rate for the legitimate user.

The optimization strategy is as follows: for the jamming \ac{o-MRAV}, its orientation is adjusted so that the null of its radiation pattern consistently points towards the legitimate user. This setup enables the jamming \ac{o-MRAV} to emit high interference power without affecting the legitimate communication. The orientation of the jamming \ac{o-MRAV} at each position is determined by this constraint. For the communicating \ac{o-MRAV}, its antenna is oriented such that the main lobe of its radiation pattern is directed towards the legitimate user, while maximizing the horizontal separation between the projected maximum antenna gain and the eavesdroppers. The orientation strategy is thus unique for each position. Finally, the 3D positions and transmission powers of both \acp{o-MRAV} are optimized.

The proposed approach is compared with two benchmark strategies: (i) \textit{Interior-point method}, where the optimization problem is solved \ac{wrt} all variables, including orientations, positions, and powers of the \acp{MRAV}, using the interior-point method; (ii) \textit{Conventional \ac{MRAV}}, assuming \acp{u-MRAV} with fixed orientations and applies the interior-point method to optimize the positions and transmission powers of the \acp{MRAV}.

\begin{figure}[tb]
    \centering
    \centering
    \adjincludegraphics[width=0.95\columnwidth, trim={{0.05\width} {0.29\height} {0.05\width} {.3\height}}, clip, keepaspectratio]{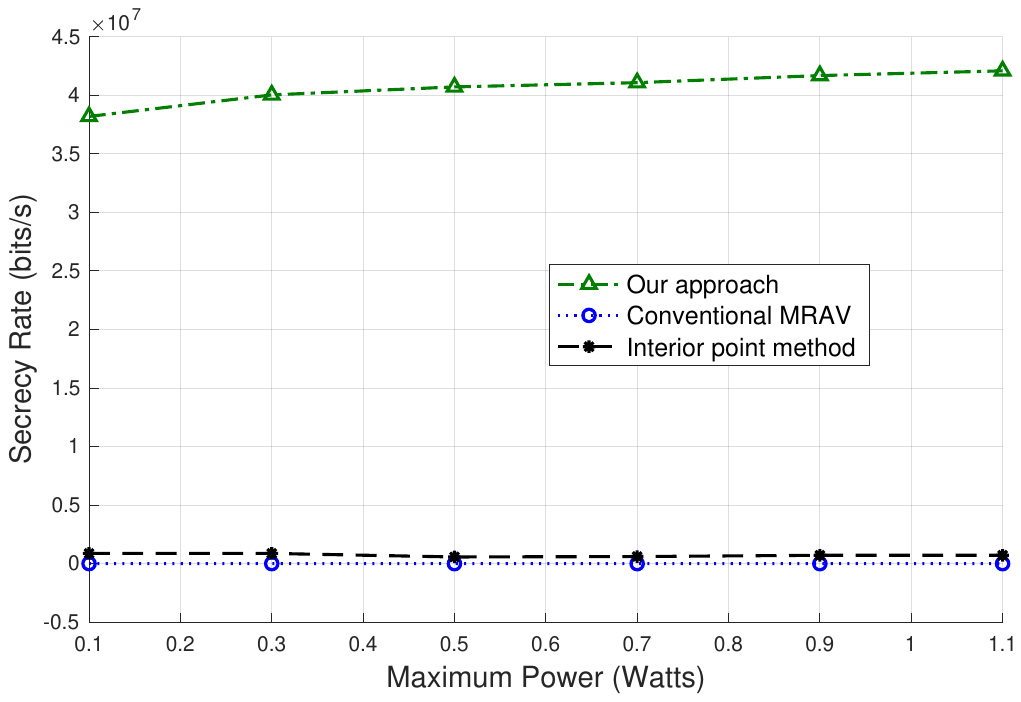}
    \caption{Secrecy rate versus maximum power for 2 eavesdroppers.}
    \label{fig4}
\end{figure}

Figure \ref{fig4} illustrates the dependency of the secrecy rate on the maximum power in case of our proposed approach, compared to the two benchmarks. The results show a significant performance advantage of the proposed approach, which provides a higher secrecy rate compared to the interior-point method. This improvement is due to the non-convex nature of the secrecy rate function, where the interior-point method can become stuck in a local optimum, leading to suboptimal performance. Furthermore, Figure \ref{fig4} shows that the conventional \ac{u-MRAV} performs the worst, as it cannot optimize its antenna orientation, leading to considerably lower performance in terms of secrecy rate. Additional information is available in \cite{Bonilla2024Globecom}.



\section{Discussion and Future Research Directions}
\label{sec:discussionFutureResearchDirections}

The introduction of \acp{o-MRAV} into communication networks offers the potential to optimize both the 3D position and 3D orientation of onboard antennas. This innovation raises several key questions for future research, as described next.

\textbf{Optimal 3D Orientation and Positioning}. A significant area of investigation involves determining the best position and antenna orientation for \acp{o-MRAV} to optimize communication performance across various use cases. This includes analyzing different scenarios to understand how precise positioning and orientation can maximize \ac{SNR}, minimize interference, and ensure reliable connectivity. While a servo-based mechanism for redirecting the antenna alone is indeed more energy-efficient, the advantage of \acp{o-MRAV} lies in their ability to seamlessly integrate position and orientation control of the entire platform. This is particularly useful in dynamic environments where both the \ac{MRAV} and surrounding communication partners are in constant motion, allowing for more stable and adaptable communication links. Although the energy efficiency of directing the entire vehicle for antenna adjustment may seem less optimal, the flexibility and robustness offered by \acp{o-MRAV} in complex and rapidly changing conditions outweigh this limitation in many use cases.

\textbf{Energy Consumption}. \acp{o-MRAV} are designed with thrust forces that sometimes counteract each other, and additional weight from actuation mechanisms (see Section \ref{sec:challengesOpenProblems}) further increases energy consumption and reduces mission endurance. The inefficiency of using full thrust to orient an antenna, along with optimizing energy usage, presents a major challenge for widespread adoption of these platforms. To enhance their viability in communication applications, significant research is needed to reduce energy consumption while maintaining high communication performance. Future work should consider integrating servo-based mechanisms for antenna redirection, which could complement \acp{o-MRAV} by offering hybrid solutions that enhance energy efficiency. The primary strength of \acp{o-MRAV} is their ability to control both position and orientation simultaneously, making them particularly useful in dynamic environments where both the \ac{MRAV} and its communication partners are constantly in motion. Another promising solution to address energy consumption is the use of tethered \acp{o-MRAV}, which can provide continuous power supply. This would be especially beneficial for applications requiring prolonged operation, such as long-term surveillance or persistent communication relay tasks.

\textbf{Trajectory Planning}. Another crucial aspect involves planning the trajectory of \acp{o-MRAV} to optimize their movements for efficient communication and navigation. This task requires careful consideration of the dynamical model of \acp{o-MRAV}, taking into account the forces and torques applied to the \ac{CoM} by the rotors, and the orientation of the antenna to enhance the transmission rate. The goal is to ensure that these vehicles can navigate complex environments while maintaining optimal positions for communication.

\textbf{2D and 3D beamforming}. It is essential to investigate the performance of \ac{o-MRAV}-assisted networks when equipped with antenna arrays. These arrays facilitate beamforming techniques, enabling precise control over beam direction. Integrating \acp{o-MRAV} with 2D and 3D beamforming has the potential to substantially enhance network performance and support higher data rates. 

These areas of research are critical for unlocking the full potential of \acp{o-MRAV} to enhance the performance of communication systems.



\section{Conclusions}
\label{sec:conclusions}

This paper introduced \aclp{o-MRAV} and outlined their capability to independently control 3D position and orientation. A comparison with \ac{u-MRAV} and \ac{f-MRAV} systems was provided, demonstrating the superior hovering ability, trajectory tracking, and rotor failure robustness of \acp{o-MRAV}. The study showcased how \acp{o-MRAV} can improve network performance in areas such as physical layer security, localization, and network capacity. Simulation results showed their effectiveness in mitigating jamming and eavesdropping attacks, emphasizing their potential in secure communications. Future research should focus on optimizing 3D orientation and positioning, minimizing energy consumption, planning efficient trajectories, and integrating advanced beamforming techniques. Our findings indicate that \acp{o-MRAV} can offer significant improvements over traditional \ac{MRAV} designs, making them a promising technology for future communication networks.



\bibliographystyle{IEEEtran}
\bibliography{references}

\begin{thebibliography}{10}
\providecommand{\url}[1]{#1}
\csname url@samestyle\endcsname
\providecommand{\newblock}{\relax}
\providecommand{\bibinfo}[2]{#2}
\providecommand{\BIBentrySTDinterwordspacing}{\spaceskip=0pt\relax}
\providecommand{\BIBentryALTinterwordstretchfactor}{4}
\providecommand{\BIBentryALTinterwordspacing}{\spaceskip=\fontdimen2\font plus
\BIBentryALTinterwordstretchfactor\fontdimen3\font minus
  \fontdimen4\font\relax}
\providecommand{\BIBforeignlanguage}[2]{{%
\expandafter\ifx\csname l@#1\endcsname\relax
\typeout{** WARNING: IEEEtran.bst: No hyphenation pattern has been}%
\typeout{** loaded for the language `#1'. Using the pattern for}%
\typeout{** the default language instead.}%
\else
\language=\csname l@#1\endcsname
\fi
#2}}
\providecommand{\BIBdecl}{\relax}
\BIBdecl

\bibitem{MozaffariIEEECST2019}
M.~Mozaffari, W.~Saad, M.~Bennis, Y.-H. Nam, and M.~Debbah, ``{A Tutorial on
  UAVs for Wireless Networks: Applications, Challenges, and Open Problems},''
  \emph{IEEE Communications Surveys \& Tutorials}, vol.~21, no.~3, pp.
  2334--2360, 2019.

\bibitem{BonillaPIEEE2024}
D.~Bonilla~Licea, M.~Ghogho, and M.~Saska, ``{When Robotics Meets Wireless
  Communications: An Introductory Tutorial},'' \emph{Proceedings of the IEEE},
  vol. 112, no.~2, pp. 140--177, 2024.

\bibitem{HamandiIJRR2021}
M.~Hamandi, F.~Usai, Q.~Sablé, N.~Staub, M.~Tognon, and A.~Franchi, ``{Design
  of multirotor aerial vehicles: A taxonomy based on input allocation},''
  \emph{The International Journal of Robotics Research}, vol.~40, no. 8--9, pp.
  1015--1044, 2021.

\bibitem{Hamandi2021AIRPHARO}
M.~Hamandi, Q.~Sable, M.~Tognon, and A.~Franchi, ``{Understanding the
  omnidirectional capability of a generic multi-rotor aerial vehicle},'' in
  \emph{Aerial Robotic Systems Physically Interacting with the Environment},
  2021, pp. 1--6.

\bibitem{Muralidharan2021ARCRAS}
A.~{Muralidharan} \emph{et~al.}, ``{Communication-Aware Robotics: Exploiting
  Motion for Communication},'' \emph{Annual Review of Control, Robotics, and
  Autonomous Systems}, vol.~4, pp. 115--139, 2021.

\bibitem{KoruASME2024ToM}
A.~T. Koru, J.~Chang, and Y.~Wan, ``{RSSI-Based Distributed Control to Align
  Directional Antenna Pairs for UAV Communication},'' \emph{IEEE Transactions
  on Mechatronics}, vol.~29, no.~4, pp. 2877--2885, 2024.

\bibitem{YuCCNC2023}
T.~Yu, K.~Kajiwara, K.~Araki, and K.~Sakaguchi, ``{Experiment of Multi-UAV
  Full-Duplex System Equipped with Directional Antennas},'' in \emph{IEEE 20th
  Consumer Communications \& Networking Conference}, 2023, pp. 1--5.

\bibitem{LiIET2019}
S.~Li, C.~He, M.~Liu, Y.~Wan, Y.~Gu, J.~Xie, S.~Fu, and K.~Lu, ``{Design and
  implementation of aerial communication using directional antennas: learning
  control in unknown communication environments},'' \emph{IET Control Theory \&
  Applications}, vol.~13, no.~17, pp. 2906--2916, 2019.

\bibitem{Aboudorra2023JINT}
Y.~{Aboudorra} \emph{et~al.}, ``{Modelling, Analysis, and Control of OmniMorph:
  an Omnidirectional Morphing Multi-rotor UAV},'' \emph{Journal of Intelligent
  \& Robotic Systems}, vol. 110, no.~21, pp. 1--14, 2024.

\bibitem{BonillaICASSP2024}
D.~Bonilla~Licea, G.~Silano, M.~Ghogho, and M.~Saska, ``{Omnidirectional
  Multi-Rotor Aerial Vehicle Pose Optimization: A Novel Approach to Physical
  Layer Security},'' in \emph{IEEE International Conference on Acoustics,
  Speech and Signal Processing}, 2024, pp. 9021--9025.

\bibitem{Bonilla2024Globecom}
D.~Bonilla~Licea, H.~El~Hammouti, G.~Silano, and M.~Saska, ``{Harnessing the
  Potential of Omnidirectional Multi-Rotor Aerial Vehicles in Cooperative
  Jamming Against Eavesdropping},'' in \emph{IEEE Global Communications
  Conference}, 2024, {To Appear}.

\bibitem{Lei2024TAES}
J.~Lei, T.~Meng, K.~Wang, W.~Wang, S.~Sun, and L.~Wang, ``{Adaptive
  Reduced-Attitude Control for Spacecraft Boresight Alignment With Safety
  Constraints and Accuracy Requirements},'' \emph{IEEE Transactions on
  Aerospace and Electronic Systems}, vol.~60, no.~4, pp. 3936--3953, 2024.

\bibitem{WangIEEETWC2022}
W.~Wang and W.~Zhang, ``{Jittering Effects Analysis and Beam Training Design
  for UAV Millimeter Wave Communications},'' \emph{IEEE Transactions on
  Wireless Communications}, vol.~21, no.~5, pp. 3131--3146, 2022.

\bibitem{Allenspach2020IJRR}
M.~Allenspach, K.~Bodie, M.~Brunner, L.~Rinsoz, Z.~Taylor, M.~Kamel,
  R.~Siegwart, and J.~Nieto, ``{Design and optimal control of a tiltrotor
  micro-aerial vehicle for efficient omnidirectional flight},'' \emph{The
  International Journal of Robotics Research}, vol.~39, no. 10--11, pp.
  1305--1325, 2020.

\bibitem{DabiriIEEETVT2019}
M.~T. Dabiri, S.~M.~S. Sadough, and I.~S. Ansari, ``{Tractable Optical Channel
  Modeling Between UAVs},'' \emph{IEEE Transactions on Vehicular Technology},
  vol.~68, no.~12, pp. 11\,543--11\,550, 2019.

\end{thebibliography}



\section{Biographies}
\label{sec:biographies}

\textbf{\textsc{Daniel Bonilla Licea}} earned his Ph.D. in 2016 from the University of Leeds. Currently, he is an assistant professor at Mohammed VI Polytechnic (UM6P) University, and an associated researcher at the Czech Technical University in Prague (CTU-P). His research focuses on signal processing and communications-aware robotics. His main research interest is on communications-aware robotics.

\textbf{\textsc{Giuseppe Silano}} is a tenured researcher at Ricerca sul Sistema Energetico and an associated researcher at CTU-P. He earned his Ph.D. in 2020 from the University of Sannio. His research interests include UAV motion planning, formal methods for robotics, and communication-aware robotics. He has authored over 25 peer-reviewed publications. 

\textbf{\textsc{Hajar Elhammouti}} joined UM6P University as an assistant professor in 2021. She earned her Ph.D. from the National Institute of Posts and Telecommunications, Rabat in 2017. Her research focuses on modeling and optimizing aerial and spatial communication systems.

\textbf{\textsc{Mounir Ghogho}} received his Ph.D. in 1997 from the National Polytechnic Institute of Toulouse. From 1997 to 2001, he was an EPSRC Research Fellow at the University of Strathclyde, and became a full Professor at the University of Leeds in 2008. In 2010, he joined the International University of Rabat, where he currently serves as Dean of the College of Doctoral Studies and Director of TICLab. He is a Fellow of IEEE, a recipient of the 2013 IBM Faculty Award and the 2000 U.K. Royal Academy of Engineering Research Fellowship. His research focuses on machine learning, signal processing, and wireless communication. He has served as AE for several prestigious journals and is a member of multiple IEEE TCs.

\textbf{\textsc{Martin Saska}} received his Ph.D. from the University of Wuerzburg in 2009. He was a Visiting Scholar at the University of Illinois at Urbana-Champaign in 2008 and the University of Pennsylvania in 2012, 2014, and 2016. Since 2009, he has been at CTU-P, where he is now an Associate Professor, leading the MRS Lab and co-founding the Center for Robotics and Autonomous Systems. He has authored over 90 peer-reviewed conference papers and 60 journal articles.


\end{document}